\documentclass[runningheads]{llncs}
\usepackage{graphicx}
\usepackage{soul}
\usepackage{color, xcolor}
\usepackage{bm}
\usepackage{amsfonts}       
\usepackage{nicefrac}       
\usepackage{microtype}      
\usepackage{amsmath}
\usepackage{amssymb}
\usepackage{booktabs}
\usepackage{url}
\usepackage{threeparttable}

\begin{document}

\title{Consistency-guided Meta-Learning for Bootstrapping Semi-Supervised Medical Image
Segmentation}

\titlerunning{MLB-Seg}

\author{Qingyue Wei\inst{1} \and
Lequan Yu\inst{2} \and
Xianhang Li\inst{3} \and
Wei Shao\inst{4} \and
Cihang Xie\inst{3} \and
Lei Xing\inst{1} \and
Yuyin Zhou\inst{3}}

\authorrunning{Q. Wei et al.}

\institute{Stanford University, Stanford, CA, USA \and The University of Hong Kong, Hong Kong, China
\and University of California, Santa Cruz, CA, USA
\and University of Florida, FL, USA
}

\maketitle             

\begin{abstract}
Medical imaging has witnessed remarkable progress but usually requires a large amount of high-quality annotated data which is time-consuming and costly to obtain. To alleviate this burden, semi-supervised learning has garnered attention as a potential solution. In this paper, we present \textbf{M}eta-\textbf{L}earning for \textbf{B}ootstrapping Medical Image \textbf{Seg}mentation (MLB-Seg), a novel method for tackling the challenge of semi-supervised medical image segmentation. Specifically, our approach first involves training a segmentation model on a small set of clean labeled images to generate initial labels for unlabeled data. To further optimize this bootstrapping process, we introduce a per-pixel weight mapping system that dynamically assigns weights to both the initialized labels and the model's own predictions. These weights are determined using a meta-process that prioritizes pixels with loss gradient directions closer to those of clean data, which is based on a small set of precisely annotated images. To facilitate the meta-learning process, we additionally introduce a consistency-based Pseudo Label Enhancement (PLE) scheme that improves the quality of the model's own predictions by ensembling predictions from various augmented versions of the same input. In order to improve the quality of the weight maps obtained through multiple augmentations of a single input, we introduce a mean teacher into the PLE scheme. This method helps to reduce noise in the weight maps and stabilize its generation process. Our extensive experimental results on public atrial and prostate segmentation datasets demonstrate that our proposed method achieves state-of-the-art results under semi-supervision.
Our code is available at \url{https://github.com/aijinrjinr/MLB-Seg}.

\end{abstract}

\section{Introduction}
Reliable and robust segmentation of medical images plays a significant role in clinical diagnosis~\cite{masood2015survey}. 
In recent years, deep learning has led to significant progress in image segmentation tasks~\cite{he2017mask,zhou2018unet++}. However, training these models~\cite{zhao2018deep} requires large-scale image data with precise pixel-wise annotations, which are usually time-consuming and costly to obtain. To address this challenge, recent studies have explored semi-supervised learning approaches for medical image segmentation, leveraging unlabeled data to enhance performance~\cite{adiga2022leveraging,xiang2022fussnet}. Semi-supervised learning (SSL) commonly uses two methods: pseudo labeling and consistency regularization. Pseudo labeling uses a model's predictions on unlabeled data to create ``pseudo labels'' and augment the original labeled data set, allowing for improved accuracy and reduced cost in training~\cite{lee2013pseudo,bachman2014learning,liu2022acpl}. Despite its potential benefits, the pseudo labeling approach may pose a risk to the accuracy of the final model by introducing inaccuracies into the training data.
Consistency regularization, on the other hand, aims to encourage the model's predictions to be consistent across different versions of the same input. UA-MT~\cite{yu2019uncertainty} and DTC~\cite{luo2020semi} are examples of consistency regularization-based methods that use teacher-student models for segmentation and emphasize dual consistency between different representations, respectively. 

Combining both methods can potentially lead to improved performance, 
as pseudo labeling leverages unlabeled data and consistency regularization improves the model's robustness~\cite{berthelot2019mixmatch,wu2021semi}. However, 
pseudo labeling still inevitably introduces inaccuracies that can hinder the model's performance.
To overcome this challenge, we propose a novel consistency-guided meta-learning framework called \textbf{M}eta-\textbf{L}earning for \textbf{B}ootstrapping Medical Image \textbf{Seg}mentation (MLB-Seg). Our approach uses pixel-wise weights to adjust the importance of each pixel in the initialized labels and pseudo labels during training. We learn these weights through a meta-process that prioritizes pixels with loss gradient direction closer to those of clean data, using a small set of clean labeled images. To further improve the quality of the pseudo labels, we introduce a consistency-based Pseudo Label Enhancement (PLE) scheme that ensembles predictions from augmented versions of the same input. To address the instability issue arising from using data augmentation, we incorporate a mean-teacher model to stabilize the weight map generation from the student meta-learning model, which leads to improved performance and network robustness. Our proposed approach has been extensively evaluated on two benchmark datasets, LA~\cite{chen2018multi} and PROMISE12~\cite{litjens2014evaluation}, and has demonstrated superior performance compared to existing methods.

\section{Method}
We present MLB-Seg, a novel consistency-guided meta-learning framework for semi-supervised medical image segmentation. Assume that we are given a training dataset consisting of clean data $D_c = \{(x^c_i, y^c_i)\}^N_{i=1}$, and unlabeled data $D_u = \{(x^u_{k})\}^K_{k=1}$ ($N \ll K$), where the input image $x^c_i,  x^u_{k}$ are of size $H \times W$ with the corresponding clean ground-truth mask $y^c_i$.

\begin{figure*}[t]
\centering
\includegraphics[width=\linewidth]{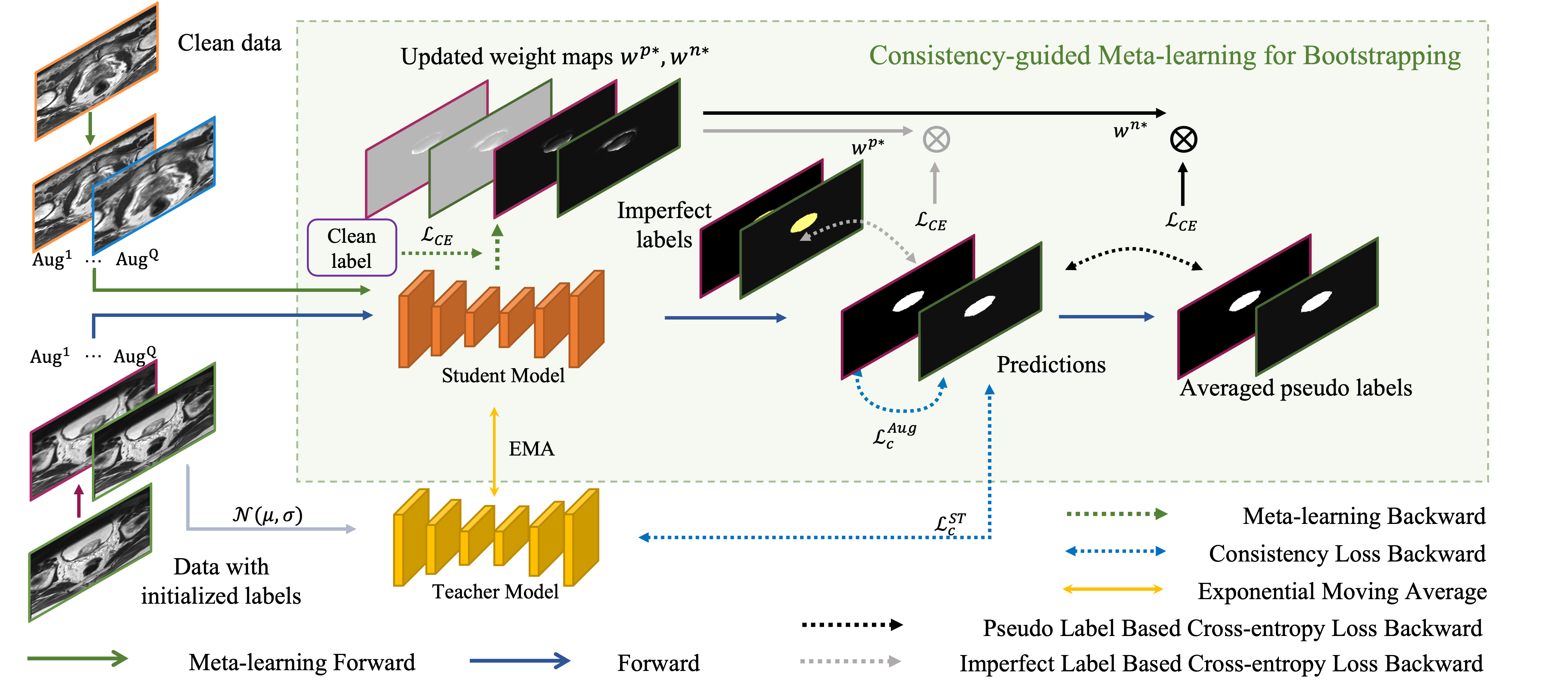}
\caption{Schematic of the proposed MLB-Seg. 
Weight maps ${w}^{n*}, {w}^{p*}$ associated with the initialized labels and pseudo labels
are meta-learned
and optimization is further improved by enhancing the pseudo label estimation. A mean teacher model is used to provide guidance for stabilizing the weight meta-update in the student model.
}
\label{fig:network sturcture}
\end{figure*}

\subsection{Meta-Learning for Bootstrapping}
\label{sec:L2B}
We first estimate labels for all unlabeled data using the baseline model which is trained on the clean data, denoted as follows
\begin{equation}\label{prediction_noisy_label}
    y^{\tilde{n}}_{k} = f_c(x^u_{k}; \theta_c),
\end{equation}
\noindent where $f_c(:;\theta_c)$ denotes the trained model parameterized by $\theta_c$ and $k=1,2,...,K$. We further denote them as $\widetilde{D_{n}}=\{(x^{\tilde{n}}_{j}, y^{\tilde{n}}_{j})\}^{K}_{j=1}$.

We then develop a novel meta-learning model for medical image segmentation, which learns from the clean set $D_c$ to bootstrap itself up by leveraging the learner’s own predictions (\emph{i.e.}, pseudo labels), called Meta-Learning for Bootstrapping (MLB). As shown in Figure~\ref{fig:network sturcture},
by adaptively adjusting the contribution between the initialized and pseudo labels commensurately in the loss function, our method effectively alleviates the negative effects from the erroneous pixels. Specifically, at training step $t$, given a training batch from $\widetilde{D_{n}}$ that $S^{n} = \{(x^{\tilde{n}}_{j}, y^{\tilde{n}}_{j}), 1 \leq j \leq b_n\}$ and a clean training batch $S^{c}=\{(x^c_{i}, y^c_{i}), 1 \leq i \leq b_c\}$ where $b_n, b_c$ are the batch size respectively. Our objective is:
\begin{equation}\label{l2b loss reweighting}
\begin{split}
    \theta^*(w^n,{w}^p) = \mathop{\arg\min}\limits_{\theta} \sum_{j=1}^{K} {w}^n_j\circ\mathcal{L}(f(x^{\tilde{n}}_{j}; \theta), y^{\tilde{n}}_{j}) + {w}^p_j\circ\mathcal{L}(f(x^{\tilde{n}}_{j}; \theta), y^{p}_{j}),
\end{split}
\end{equation}
\begin{equation}\label{pseudo label generation}
\begin{split}
y^{p}_{j} = \mathop{\arg\max}\limits_{c=1,...,C}f(x^{\tilde{n}}_{j}; \theta),
\end{split}
\end{equation}
\noindent where $y^{p}_{j}$ is the pseudo label  generated by $f(x^{\tilde{n}}_{j}; \theta)$, $\mathcal{L}(\cdot)$ is the cross-entropy loss function, $C$ is the number of classes (we set $C=2$ throughout this paper), ${w}^n_j, {w}^p_j \in \mathbb{R}^{H \times W}$ are the weight maps used for adjusting the contribution between the initialized and the pseudo labels in two different loss terms. $\circ$ denotes the Hadamard product. 
We aim to solve for Eq.~\ref{l2b loss reweighting}  following 3 steps:
\begin{itemize}
\item \textbf{Step 1:} Update $\hat{\theta}_{t+1}$ based on $S^{n}$ and current weight map set. 
Following~\cite{ren2018learning}, 
during training step $t$, we calculate $\hat{\theta}_{t+1}$ to approach the optimal $\theta^*({w}^n, {w}^p)$ as follows:
\begin{equation}\label{L2Bstep1}
\begin{split}
    \hat{\theta}_{t+1} = \theta_t - \alpha\nabla(\sum_{j=1}^{b_n}{w}^n_j\circ\mathcal{L}(f(x^{\tilde{n}}_{j}; \theta), y^{\tilde{n}}_{j}) + {w}^p_j\circ\mathcal{L}(f(x^{\tilde{n}}_{j}; \theta), y^{p}_{j}))\big|_{\theta=\theta_t},
\end{split}
\end{equation}
\noindent where $\alpha$ represents the step size.

\item \textbf{Step 2:}  Generate the meta-learned weight maps ${w}^{n*},{w}^{p*}$ based on $S^{c}$ and $\hat{\theta}_{t+1}$  
by minimizing the standard cross-entropy loss in
the meta-objective function over the clean training data:
\begin{equation}\label{weight map updated}
\begin{split}
    {w}^{n*}, {w}^{p*}  =  \mathop{\arg\min}\limits_{{w}^n, {w}^p \geq \boldsymbol{0}} \frac{1}{N}\sum_{i=1}^N \mathcal{L}(f(x^c_{i}; \theta^*({w}^n,{w}^p)), y^c_{i}).
\end{split}
\end{equation}
Note that here we restrict every element in ${w}^{n/p}$ to be non-negative to prevent potentially unstable training~\cite{ren2018learning}. 
Such a meta-learned process yields weight maps which can better balance the contribution of the initialized and the pseudo labels, thus reducing the negative effects brought by the erroneous pixels. 
Following~\cite{ren2018learning,zhou2022learning}, we only apply one-step gradient descent of ${w}^{n/p}_j$ based on a small clean-label data set $S^c$, to reduce the computational expense. 
To be specific, at training step $t$, ${w}^{n/p}_j$ is first initialized as $\boldsymbol{0}$, then we estimate ${w}^{n*}, {w}^{p*}$ as: 
\begin{equation}\label{L2Bstep2}
\begin{split}
   ({w}^n_{j,t}, {w}^p_{j,t}) = -\beta \nabla(\frac{1}{b_c}\sum_{i=1}^{b_c} \mathcal{L}(f(x^c_{i}; \hat{\theta}_{t+1}), y^c_{i}))\big|_{{w}^n_j, {w}^p_j=\boldsymbol{0}},
\end{split}
\end{equation}
\begin{equation}\label{L2Bstep2_nonneg}
\begin{split}
    \widetilde{{w}}^{{n}_{r,s}}_{j,t} = \text{max}({w}^{{n}_{r,s}}_{j,t}, 0),\quad \widetilde{{w}}^{{p}_{r,s}}_{j,t} = \text{max}({w}^{{p}_{r,s}}_{j,t}, 0),
\end{split}
\end{equation}
\begin{equation}\label{L2Bstep2_norm}
\begin{split}
    \widetilde{{w}}^{{n}_{r,s}}_{j,t} = \frac{\widetilde{{w}}^{{n}_{r,s}}_{j,t}}{\sum_{j=1}^{b_n}\sum\limits_{r,s}\widetilde{{w}}^{{n}_{r,s}}_{j,t} + \epsilon}, \widetilde{{w}}^{{p}_{r,s}}_{j,t} = \frac{\widetilde{{w}}^{{p}_{r,s}}_{j,t}}{\sum_{j=1}^{b_n}\sum\limits_{r,s}\widetilde{{w}}^{{p}_{r,s}}_{j,t} + \epsilon},
\end{split}
\end{equation}
\noindent where $\beta$ stands for the step size and ${{w}}^{{n}/p_{r,s}}_{j,t}$ indicates the value at $r^{th}$ row, $s^{th}$ column of ${{w}}^{{n}/p}_{j}$ at time $t$. Eq. \ref{L2Bstep2_nonneg} is used to enforce all weights to be strictly non-negative. 
Then Eq. \ref{L2Bstep2_norm} is introduced to normalize the weights in a single training
batch so that they sum up to one. Here, we add a small number $\epsilon$ to keep the denominator greater than 0.
\item \textbf{Step 3:} The meta-learned weight maps are used to spatially modulate the pixel-wise loss to update $\theta_{t+1}$: 
\begin{equation}\label{L2Bstep3}
\begin{split}
    \theta_{t+1} = \theta_t - \alpha\nabla{(\sum_{j=1}^{b_n} \widetilde{{w}}^n_{j,t}\circ\mathcal{L}(f(x^{\tilde{n}}_{j}; \theta), y^{\tilde{n}}_{j}) + \widetilde{{w}}^p_{j,t}\circ\mathcal{L}(f(x^{\tilde{n}}_{j}; \theta), y^{p}_{j}))}\big|_{\theta=\theta_t}.
\end{split}
\end{equation}
\end{itemize}

\subsection{Consistency-based Pseudo Label Enhancement}
\label{sec:PLE}
To generate more reliable pseudo labels,
we propose Pseudo Label Enhancement (PLE) scheme based on consistency, which enforces consistency across augmented versions of the same input.
Specifically, we perform $Q$ augmentations on the same input image and enhance the pseudo label by averaging the outputs of the $Q$ augmented versions and the original input:
\begin{equation}\label{Batch_Augmentation_generate}
\begin{split}
   \widehat{y}^{p}_{j} = \mathop{\arg\max}\limits_{c=1,...,C}\frac{1}{Q+1}(\sum^{Q}_{q=1}\tau^{-1}_q(f(x^{{\tilde{n}}^{q}}_{j}; \theta)) + f(x^{{\tilde{n}}^{0}}_{j}; \theta)),
\end{split}
\end{equation}
\noindent where $f(x^{{\tilde{n}}^{q}}_{j}; \theta)$ is the output of $q$-th augmented sample, $f(x^{{\tilde{n}}^{0}}_{N+j}; \theta)$ is the output of the original input, and $\tau^{-1}_q$ means the corresponding inverse transformation of the $q$-th augmented sample. 
Meanwhile, to further increase the output consistency among all the augmented samples and original input, we introduce an additional consistency loss $\mathcal{L}^{Aug}_{c}$ to the learning objective:
\begin{equation}\label{Batch_Augmentation_consistency_loss}
\begin{split}
   \mathcal{L}^{Aug}_{c}(x^{\tilde{n}}_{j}) = \frac{2}{(Q+1)Q}\frac{1}{HW}\sum_{q,v}\sum_{r,s}||f(x^{{\tilde{n}}^{q}}_{j}; \theta)_{r,s}  - \tau_q(\tau^{-1}_v(f(x^{{\tilde{n}}^{v}}_{j}; \theta)))_{r,s} ||^2,
\end{split}
\end{equation} 
\noindent where $(r,s)$ denotes the pixel index. $\tau_q$ is the corresponding transformation to generate the $q$-th augmented sample. $(q,v)$ denotes the pairwise combination among all augmented samples and the original input. The final loss is the mean square distance among all $\frac{(Q+1)Q}{2}$ pairs of combinations.

\subsection{Mean Teacher for Stabilizing Meta-Learned Weights}
Using PLE alone can result in performance degradation with increasing numbers of augmentations due to increased noise in weight maps. This instability can compound during subsequent training iterations. To address this issue during meta-learning, we propose using a mean teacher model~\cite{tarvainen2017mean} with consistency loss(Eq.~\ref{Mean_Teacher_L_c}). The teacher network guides the student meta-learning model, stabilizing weight updates and resulting in more reliable weight maps. Combining PLE with the mean teacher model improves output robustness. The student model is used for meta-learning, while the teacher model is used for weight ensemble with Exponential Moving Average (EMA)~\cite{tarvainen2017mean} applied to update it. The consistency loss maximizes the similarity between the teacher and student model outputs, adding reliability to the student model and stabilizing the teacher model. For each input $x^{\tilde{n}}_{j}$ in the batch $S^n$, we apply disturbance to the student model input to become the teacher model input. Then a consistency loss $\mathcal{L}^{ST}_c$ is used to maximize the similarity between the outputs from the teacher model and student model, further increasing the student model's reliability while stabilizing the teacher model. Specifically, for each input $x^{\tilde{n}}_{j}$ in the batch $S^n$, then corresponding input of teacher model is
\begin{equation}\label{Teacher_model_input}
\begin{split}
    x^T_{j} = x^{\tilde{n}}_{j} + \gamma \mathcal{N}(\mu, \sigma),
\end{split}
\end{equation}
\noindent where $\mathcal{N}(\mu, \sigma)\in \mathbb{R}^{H \times W}$ denotes the Gaussian distribution with $\mu$ as mean and $\sigma$ as standard deviation. And $\gamma$ is used to control the noise level. The consistency loss is implemented based on pixel-wise mean squared error (MSE) loss:
\begin{equation}\label{Mean_Teacher_L_c}
\begin{split}
    \mathcal{L}^{ST}_c(x^{\tilde{n}}_{j}) = \frac{1}{HW}\sum_{r,s}||f(x^{\tilde{n}}_{j}; \theta^S_t)_{r,s} -  f(x^T_{j}; \theta^T_t)_{r,s}||^2.
\end{split}
\end{equation}
\noindent

\section{Experiments}
\subsection{Experimental Setup}\label{Experiments_details}
\paragraph{Datasets.}
We evaluate our proposed method on two different datasets including 1) the left atrial (LA) dataset from the 2018 Atrial Segmentation Challenge~\cite{chen2018multi} as well as 2) the Prostate MR Image Segmentation 2012 (PROMISE2012) dataset~\cite{litjens2014evaluation}. Specifically, LA dataset is split into 80 scans for training and 20 scans for evaluation following~\cite{yu2019uncertainty}. From the training set, 8 scans are randomly selected as the meta set. All 2D input images were resized to $144 \times 144$. For PROMISE2012, we randomly split 40/4/6 cases for training  and 10 for evaluation (4 for validation, 6 for test) following~\cite{peng2021self}. We randomly pick 3 cases from the training set as the meta set and resize all images to $144 \times 144$. We evaluate our segmentation performances using four metrics: the Dice coefficient, Jaccard Index (JI), Hausdorff Distance (HD), and Average Surface Distance (ASD).

\paragraph{Implementation Details.}
All of our experiments are based on 2D images. We adopt UNet++ as our baseline. Network parameters are optimized by SGD setting the learning rate at $0.005$, momentum to be $0.9$ and weight decay as $0.0005$. The exponential moving average (EMA) decay rate is set as 0.99 following~\cite{tarvainen2017mean}. For the label generation process, we first train with all clean labeled data for 30 epochs with batch size set as 16. We then use the latest model to generate labels for unlabeled data. Next, we train our MLB-Seg for 100 epochs. 

\subsection{Results under Semi-Supervision}
To illustrate the effectiveness of MLB-Seg under semi-supervision. We compare our method with the baseline (UNet++\cite{zhou2018unet++}) and previous semi-supervised methods on LA dataset (Table~\ref{compare_semi_LA}) and PROMISE12 (Table~\ref{compare_semi_PROMISE}), including an adversarial learning method~\cite{zheng2019semi}, consistency based methods~\cite{yu2019uncertainty,hang2020local,wang2020double,luo2020semi,wu2021semi}, an uncertainty-based strategy~\cite{adiga2022leveraging}, 
and contrastive learning based methods~\cite{peng2021self,wu2022exploring}. For a fair comparison in Table~\ref{compare_semi_PROMISE}, we also use UNet~\cite{ronneberger2015u} as the backbone and resize images to $256 \times 256$, strictly following the settings in Self-Paced~\cite{peng2021self}.
We split the evaluation set (10 cases) into 4 cases as the validation set, and 6 as the test set. We then select the best checkpoint based on the validation set and report the results on the test set. As shown in Table~\ref{compare_semi_LA} \& \ref{compare_semi_PROMISE}, our MLB-Seg outperforms recent state-of-the-art methods on both PROMISE12 (under different combinations of backbones and image sizes) and the LA dataset across almost all evaluation measures. 

\begin{table}[t!]
\footnotesize 
\centering
\caption{Comparison with existing methods under semi-supervision on LA dataset.}
\resizebox{0.75\columnwidth}{19mm}{
\begin{tabular}{c|c|c|c|c} \toprule
Method & Dice (\%)$\uparrow$ & JI (\%)$\uparrow$  & HD (voxel)$\downarrow$& ASD (voxel)$\downarrow$  \\ \hline
UNet++~\cite{zhou2018unet++} & 81.33 &70.87 &14.79 &4.62 \\
DAP~\cite{zheng2019semi}& 81.89 & 71.23 &15.81 &3.80 \\
UA-MT~\cite{yu2019uncertainty} & 84.25 &73.48 &13.84 &3.36 \\
SASSNet~\cite{li2020shape} &87.32 &77.72 &9.62 &2.55 \\
LG-ER-MT~\cite{hang2020local} &85.54 &75.12 &13.29 &3.77 \\
DUWM~\cite{wang2020double} &85.91 &75.75 &12.67 &3.31 \\
DTC~\cite{luo2020semi} &86.57  & 76.55 &14.47 &3.74 \\
MC-Net~\cite{wu2021semi} &87.71 &78.31 &9.36 &$\bm{2.18}$ \\
Uncertainty-Based~\cite{adiga2022leveraging} &86.58 &76.34 &11.82 & -\\
\textbf{MLB-Seg} &$\bm{88.69}$ &$\bm{79.86}$ &$\bm{8.99}$ &2.61 \\ 
\bottomrule
\end{tabular}}
\label{compare_semi_LA}
\end{table}

\begin{table}[htb!]
\footnotesize 
\centering
\caption{Comparison with existing methods under semi-supervision on PROMISE12. All methods included in the comparison have been re-implemented using our data split to ensure a fair evaluation, with the provided source codes.}
\resizebox{0.4\linewidth}{19mm}{
\begin{tabular}{c|c} \toprule%[0.05em]
Method & Dice (\%)$\uparrow$ \\ \hline
UNet++~\cite{zhou2018unet++} & 68.85\\
UA-MT~\cite{yu2019uncertainty} &65.05 \\
DTC~\cite{luo2020semi} &63.44 \\
SASSNet~\cite{li2020shape} &73.43 \\
MC-Net~\cite{wu2021semi} &72.66 \\
SS-Net~\cite{wu2022exploring} &73.19\\
Self-Paced~\cite{peng2021self} (UNet, 256) & 74.02 \\
\textbf{MLB-Seg} (UNet, 144) &76.41 \\
\textbf{MLB-Seg} (UNet, 256) &76.15 \\
\textbf{MLB-Seg} (UNet++, 144) &77.22 \\
\textbf{MLB-Seg} (UNet++, 256) &\textbf{78.27} \\ \bottomrule%[0.05em]
\end{tabular}}
\label{compare_semi_PROMISE}
\end{table}

\subsection{Ablation Study}
\label{abl}To explore how different components of our MLB-Seg contribute to the final result, we conduct the following experiments under semi-supervision on PROMISE12: 1) the bootstrapping method~\cite{reed2014training} (using fixed weights without applying meta-learning);
2)  \textbf{MLB}, which only reweights the initialized labels and pseudo labels without applying PLE and mean teacher; 3) \textbf{MLB + mean teacher} which combines MLB with mean teacher scheme; 4) \textbf{MLB + PLE} which applies PLE strategy with MLB. When applying multiple data augmentations (\emph{i.e.}, for $Q=2, 4$), we find the best performing combinations are 2 $\times$ PLE (using one zoom in and one zoom out), 4 $\times$ PLE (using one zoom in and two zoom out and one flip for each input);
5) \textbf{MLB + PLE + mean teacher} which combines PLE, mean teacher with MLB simultaneously to help better understand how mean teacher will contribute to PLE. 
Our results are summarized in Table~\ref{Ab_new}, which shows the effectiveness of our proposed components. The best results are achieved when all components are used.

To demonstrate how PLE combined with the mean teacher model help stabilize the meta-weight update, we compare the performance of MLB + PLE (w/ mean teacher) with MLB + PLE + mean teacher under different augmentations ($Q$) on PROMISE12 dataset (See supplementary materials for details). We find out that for MLB + PLE (w/o mean teacher), performance improves from 74.34\% to 74.99\% when  $Q$ is increased from 1 to 2, but decreases significantly when $Q\geq4$. Specifically, when $Q$ reaches 4 and 6, the performance significant drops from 74.99\% to 72.07\% ($Q=4$) and from 74.99\% to 70.91\% ($Q=6$) respectively. We hypothesize that this is due to increased noise from initialized labels in some training samples, which can lead to instability in weight updates. To address this issue,  we introduce the mean-teacher model~\cite{tarvainen2017mean} into PLE to stabilize weight map generation from the student meta-learning model. And MLB + PLE + mean teacher turns out to consistently improve the stability of meta-learning compared with its counterpart without using mean teacher, further validating the effectiveness of our method (see Supplementary for more examples).
Specifically, for MLB + PLE + mean teacher, the performance reaches $76.63\%$ (from $72.07\%$) when $Q=4$, $75.84\%$ (from $70.91\%$) when $Q=6$. 
\begin{table}[t]
\centering
\caption{Ablation study on different components used in MLB-Seg based on PROMISE12.}
\resizebox{0.9\linewidth}{!}{
\begin{tabular}{cccc|c|c|c|c}
\toprule
MLB & mean teacher &$2 \times$ PLE &$4\times$ PLE  & Dice  (\%)$\uparrow$ & JI (\%)$\uparrow$ & HD (\text{voxel})$\downarrow$& ASD (\text{voxel})$\downarrow$ \\ \hline
& & &  &70.85 & 55.85&10.02 &4.35 \\ 
\checkmark& & &  & 73.97   &59.71	&8.49 &3.49 \\ 
\checkmark&\checkmark & &   &73.76 &59.37	&8.04 &3.00 \\ 
\checkmark& &\checkmark &  &75.01  &60.92	&$\bm{7.58}$ &2.70 \\
\checkmark& & &\checkmark  &73.10  &58.45	&9.42 &3.57 \\
\checkmark&\checkmark & &\checkmark & $\bm{76.68}$ &$\bm{63.14}$ &7.85 &$\bm{2.64}$\\
\bottomrule
\end{tabular}}
\label{Ab_new}
\end{table}

\begin{figure}[htb]
\centering
\includegraphics[width=.8\linewidth]{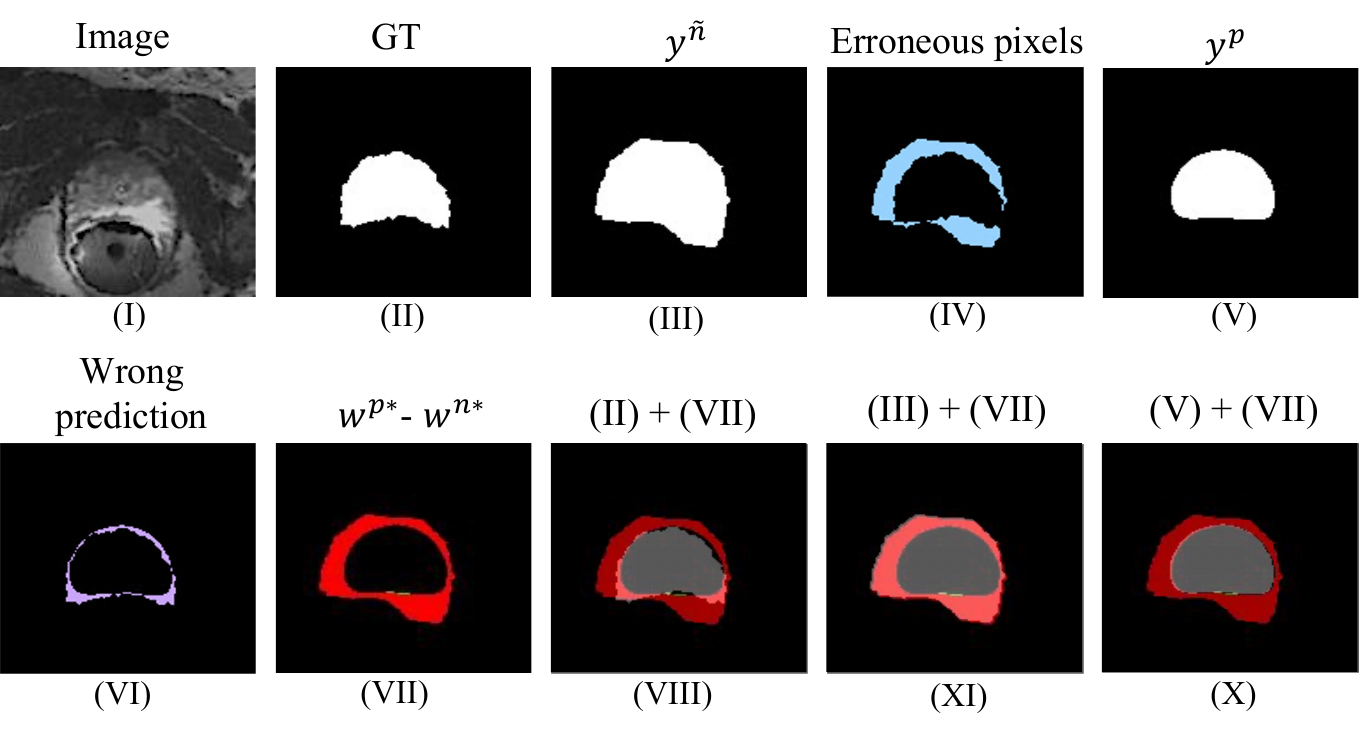}
\caption{Weight map visualization. 
}
\label{fig:weights vis}
\end{figure}

\noindent\textbf{Qualitative Analysis.} 
To illustrate the benefits of MLB-Seg for medical image segmentation, we provide a set of qualitative examples in Fig.~\ref{fig:weights vis}. In the visualization of weight maps of Fig.~\ref{fig:weights vis}, the blue/purple represents for the initialized label in $y^{\tilde{n}}$/ $y^{p}$, while the red indicates pixels in $w^{p*}$ have higher values. We observe that MLB-Seg places greater emphasis on edge information. It is evident that higher weights are allotted to accurately predicted pseudo-labeled pixels that were initially mislabeled, which effectively alleviates the negative effects from erroneously initialized labels.

\section{Conclusion}
In this paper, we propose a novel meta-learning based segmentation method for medical image segmentation under semi-supervision. With few expert-level labels as guidance, our model bootstraps itself up by dynamically reweighting the contributions from initialized labels and its own outputs, thereby alleviating the negative effects of the erroneous voxels. 
In addition, we address an instability issue arising from the use of data augmentation by introducing a mean teacher model to stabilize the weights. Extensive experiments demonstrate the effectiveness and robustness of our method under semi-supervision. Notably, our approach achieves state-of-the-art results on both the LA and PROMISE12 benchmarks. \\

\noindent\textbf{Acknowledgment.} This work is supported by the Stanford 2022 HAI Seed Grant and National Institutes of Health 1R01CA256890 and 1R01CA275772.

\bibliographystyle{splncs04}
\bibliography{paper}
\end{document}